\documentclass{svproc}
\usepackage{graphicx}%
\usepackage{multirow}%
\usepackage{amsmath,amssymb,amsfonts}%
\usepackage{mathrsfs}%
\usepackage[title]{appendix}%
\usepackage{xcolor}%
\usepackage{textcomp}%
\usepackage{manyfoot}%
\usepackage{booktabs}%
\usepackage{algorithm}%
\usepackage{algorithmicx}%
\usepackage{algpseudocode}%
\usepackage{listings}%
\usepackage{url}

\def\orcidID#1{\unskip$^{[#1]}$} 

\begin{document}
\mainmatter              
\title{Spectral Theory for Edge Pruning in Asynchronous Recurrent Graph Neural Networks}
\titlerunning{Spectral Theory for Edge Pruning in ARGNNs}  
%
\author{Nicolas Bessone\orcidID{0009-0008-9008-0003}}
\authorrunning{Nicolas Bessone} 

\institute{IT University of Copenhagen, Rued Langgaards Vej 7, 2300 København,\\
\email{nbes@itu.dk}\\
Copenhagen, Denmark}

\maketitle              

\begin{abstract}
Graph Neural Networks (GNNs) have emerged as a powerful tool for learning on graph-structured data, finding applications in numerous domains including social network analysis and molecular biology. Within this broad category, Asynchronous Recurrent Graph Neural Networks (ARGNNs) stand out for their ability to capture complex dependencies in dynamic graphs, resembling living organisms' intricate and adaptive nature. However, their complexity often leads to large and computationally expensive models. Therefore, pruning unnecessary edges becomes crucial for enhancing efficiency without significantly compromising performance. This paper presents a dynamic pruning method based on graph spectral theory, leveraging the imaginary component of the eigenvalues of the network graph's Laplacian.
\end{abstract}
\section{Introduction}

Modern deep neural networks are characterized by enormous model sizes, requiring considerable computational power and storage resources. To mitigate these demands, network pruning techniques are frequently used to decrease memory usage and shorten training time.

The Lottery Ticket Hypothesis \cite{frankle_lottery_2019} suggests that within a randomly initialized dense network, a sparse sub-network can be trained using the original weights to achieve performance comparable to that of the full network. Building on this idea, pruning network parameters has become a popular compression technique, operating on the premise that many parameters are often redundant and can be removed without significantly impacting the network's performance. However, identifying these redundant connections (weights) is challenging due to the complexity of the network \cite{chatzikonstantinou_recurrent_2021,liang_pruning_2021}, where millions of mathematical operations are performed, and the contributions of all weights are combined to generate the final output.

Dynamic pruning \cite{tang_manifold_2021} involves adjusting the network's topology in response to changing conditions. In a decentralized and asynchronous setting, these changes are difficult to track and respond to in real time. The network must adapt without centralized coordination, which can lead to issues such as network partitioning or inefficiencies if pruning decisions are not well-coordinated.

In a decentralized network, where there is no central authority or single point of control, each node operates independently and has only partial information about the overall network state and its current performance. This makes it difficult to implement and coordinate pruning decisions consistently across the network. Each node must make local decisions based on limited information, which can lead to inconsistencies, where some nodes might prune connections that other nodes still rely on. In an asynchronous network, nodes do not operate under a global clock. Messages between nodes can be delayed, lost, or arrive out of order. This lack of synchronization exacerbates the difficulty of dynamic pruning, as nodes might make pruning decisions based on outdated or incomplete information.

This paper introduces a dynamic pruning methodology grounded in spectral theory, a branch of mathematics that focuses on the study of the spectrum —eigenvalues and eigenvectors— of operators, particularly linear operators on vector spaces, often within the context of functional analysis. Although the literature on pruning methodologies is extensive and well-documented \cite{cheng_survey_2023,liang_pruning_2021,zhong_where_2018}, the application of spectral theory to enhance pruning techniques appears, to the best of my knowledge, to be unexplored.

This work represents a preliminary exploration into the potential of spectral theory to identify and capture unnecessary connections within a network during training. It aims to initiate a discussion on spectral theory's broader applications and implications in enhancing pruning techniques. By leveraging the principles of spectral analysis, this study seeks to uncover new avenues for optimizing neural network architectures, contributing to the ongoing research on advanced pruning methodologies.

\section{A Fully Decentralized Asynchronous Recurrent Graph Neural Network}
Asynchronous Recurrent Graph Neural Networks (ARGNNs) represent a subclass of RGNNs where the nodes update their states asynchronously. This asynchronous updating mechanism more closely mimics many real-world systems where entities do not act in a synchronized manner \cite{faber_asynchronous_2022}. The challenge with ARGNNs lies in their potential computational inefficiency and complexity due to the large number of edges that need to be processed at each time step \cite{chatzikonstantinou_recurrent_2021}.

ARGNNs operate effectively in decentralized networks where nodes lack global knowledge. In these settings, computations rely solely on local information within each node's neighborhood. This decentralized approach mirrors the adaptive behavior observed in living organisms, allowing GNNs to capture complex dependencies in dynamic graphs without requiring global coordination or oversight \cite{hemmer_optimal_2024}. This characteristic is particularly advantageous in scenarios such as social network analysis and molecular biology, where graphs evolve dynamically and maintaining global knowledge is impractical or infeasible.

In this work, a decentralized implementation of a Neural Network (NN) was utilized. During each update step, the nodes execute their behaviors in an arbitrary order, this behavior consists in updating its value according to an activation function, followed by a parameters optimization based on a local gradient descent. The activation function $f$ of each node $n$ is computed based on the aggregation of its inputs and their respective weights. This forward computation is expressed as:

\begin{equation}
    f(n) = V_{n} =\frac{1}{1 + e^{-\left(\overline{V_{n_{I}}} \cdot \overline{W_{n_{I}}} \right)}}
\label{eq:forward}
\end{equation}

\noindent where $\overline{V_{n_{I}}}$ is the vector of input values, $\overline{W_{n_{I}}}$ is the vector of edge weights to the input nodes of $n$. Following its activation, the node will compute its local error gradient $\nabla e_{n}$:

\begin{equation}
    \nabla e_{n} = \left(\overline{\nabla e_{n_{O}}} \cdot \overline{W_{n_{O}}} \right) \cdot f'(n) 
\label{eq:gradient}
\end{equation}

\noindent where $\overline{\nabla e_{n_{O}}}$ is the vector of error gradients of output nodes of $n$ while $\overline{W_{n_{O}}}$ is the vector of edge weights to the output nodes of $n$, and $f'(n)$ is the derivative of the activation function (Eq. \ref{eq:forward}), evaluated at its current state. In the case of the output node (the one utilized to evaluate the performance of the network), the gradient error is computed as the difference between the expected value and the current value $V$ of the node. This output node is the only node that introduces an error signal in the system, which is later propagated through the network as indicated in Eq. \ref{eq:gradient}, this local gradient of error is finally used in a gradient descent operation to adjust the weights of its input nodes edges $\overline{W_{n_{I}}}$:

\begin{equation}
\overline{W_{n_{I}}} = \overline{W_{n_{I}}} - \nabla e_{n} \cdot \overline{V_{n_{I}}}
\label{eq:optimization}
\end{equation}

\section{Graph Spectral Theory and Pruning}
Graph spectral theory provides a robust mathematical framework for analyzing the structural properties of graphs. The eigenvalues and eigenvectors of the graph Laplacian matrix, which encodes the connectivity information of the graph, play a pivotal role in various graph algorithms \cite{tsitsulin_graph_2023,mesbahi_graph_2010}. These eigenvalues and eigenvectors encapsulate important properties such as coupling, oscillation frequencies, connectivity, clustering tendencies \cite{jia_latest_2014}, and the presence of bottlenecks within the graph \cite{mesbahi_graph_2010}. The novel aspect of the presented pruning method focuses on the imaginary part of the eigenvalues of the Laplacian matrix. Unlike the real parts, which are often associated with the global structural properties of the graph, the imaginary parts can reveal subtle patterns related to edge redundancy and community structure.

The pruning strategy leverages spectral graph theory, which helps capture important structural and dynamic properties of the network through the eigenvalues of its local Laplacian matrix. Each node computes the Laplacian matrix based on its local connections, and the eigenvalues of this matrix give insight into the node's interactions with its neighbors.

The key insight here is the relationship between eigenvalues with opposite imaginary parts. Imaginary components in the eigenvalues typically arise when there is oscillatory or cyclic behavior in the local network, often indicating some form of dynamic coupling between the nodes. When two nodes in a graph have eigenvalues with equal magnitudes but opposite imaginary components, it suggests that their behaviors are symmetrically coupled—meaning the actions or states of one node have a mirrored impact on the other.

This symmetry implies redundancy: if the interaction between two nodes can be described by symmetric dynamics, one of these interactions can often be pruned without significantly altering the overall behavior of the system. Specifically, one of the weights can be expressed as a linear transformation of the other. Furthermore, if the dynamics between two nodes are independent, their corresponding eigenvalues will be zero. By removing the edge between nodes with opposite imaginary eigenvalues, we reduce the number of trainable parameters while preserving the essential properties of the graph.

The use of eigenvalues as a pruning criterion ensures that the graph retains its core dynamic and structural characteristics, even as unnecessary edges are removed. This selective pruning balances model complexity and expressiveness, enabling efficient training without sacrificing important relationships within the network.

\subsection{Degree Matrix and Weighted Degree Matrix}
Before delving into the pruning algorithm, it is essential to highlight a few important points regarding the Degree matrix. Each node's pruning decision is based on the eigenvalues of the Laplacian matrix corresponding to its neighborhood, expressed as follows:

\begin{equation}
L_{n} = D_{n} - A_{n} \odot W_{n}
\label{eq:laplacian}
\end{equation}

\noindent being $D_{n}$ the local Degree matrix, $A_{n}$ the neighborhood adjacency matrix, and $W_{n}$ the corresponding weight matrix of the node $n$, $\odot$ represents the Hadamard product (element-wise product).

The Degree matrix $D_{n}$ captures the number of edges in the node's neighborhood. In an undirected graph, the degree matrix $D$ is a diagonal matrix where each diagonal entry $D_{ii}$ represents the degree of the node $i$, which is the number of edges connected to it. In this case, $D$ captures the number of connections each node has.

In directed graphs, the concept of degree is split into in-degrees and out-degrees: $D_{in}$ is a diagonal matrix where each diagonal entry $D_{ii}$ represents the in-degree of node $i$, and $D_{out}$ is a diagonal matrix where each diagonal entry $D_{ii}$ represents the out-degree of node $i$. The degree matrix $D$ for a directed graph is computed as $D = D_{in} + D_{out}$. Here, $D$ combines both incoming and outgoing connections, capturing the total degree in terms of both input and output edges.

In weighted graphs, each edge has an associated weight. In this setting, the degree matrices need to account for these weights: in $D_{Win}$, for each node $i$, the diagonal entry $D_{ii}$ is the sum of the weights of all edges directed towards node $i$, and in $D_{Wout}$, for each node $i$, the diagonal entry $D_{ii}$ is the sum of the weights of all edges directed away from node $i$. In this context, the total degree matrix is given by: $D_{W} = D_{Win} + D_{Wout}$.

The Laplacian matrix is a fundamental construct that reveals important properties about the structure of a graph. In the context of optimizing edge weights within an ARGNN and identifying redundant couplings, our goal is to construct a Laplacian matrix that effectively represents the relationships between these edge weights. We will assess the performance of our pruning strategy using two types of Degree matrix: the directed Degree matrix $D = D_{in} + D_{out}$ and the weighted Degree matrix $D_{W} = D_{W_{in}} + D_{W_{out}}$. The computation of the Degree matrix $D_{W}$ may result in negative or zero values on its diagonal for certain weight configurations. In contrast, the directed Degree matrix $D$ ensures positive values on its diagonal, with zero indicating an isolated node. The motivation for exploring both configurations of $D$ is to explore different ways to construct the Laplacian matrix that may result in different pruning performances.

\subsection{Pruning Decision}
After the parametric optimization shown in Eq. \ref{eq:optimization}, each node will compute the Laplacian matrix of its neighborhood expressed in Eq. \ref{eq:laplacian}.
\noindent Let $\overline{\lambda_{n}}$ be the vector of eigenvalues of the local Laplacian matrix of the node $n$:
\begin{equation}
   \overline{\lambda_{n}} = \text{eig}(L_{n}) = [\lambda_0, \lambda_1, \dots, \lambda_m]
\end{equation}

\noindent Compute the imaginary parts of these eigenvalues as:
\begin{equation}
   \Im(\overline{\lambda_{n}}) = [\Im(\lambda_0), \Im(\lambda_1), \dots, \Im(\lambda_m)]
\end{equation}

\noindent Let $a$ be defined as the maximum imaginary component among the eigenvalues:
\begin{equation}
   a = \max\left(\Im(\overline{\lambda_{n}})\right) > 0
   \end{equation}

Suppose there exists another eigenvalue $\lambda_k$ with an imaginary component $b$ such that $b = -a$. Let $n_a$ and $n_b$ be the nodes associated with the eigenvalues having imaginary components $a$ and $b$, respectively. The edge between the nodes $n_a$ and $n_b$, specifically the directed edge $n_a \leftarrow n_b$, is removed if it exists. Implementation and reproduction kit available at \cite{bessone_nhbessspectral-theory-for-edge-pruning--argnn_2024}. 

\begin{figure}[t]
    \centering
    \includegraphics[width=0.4\linewidth]{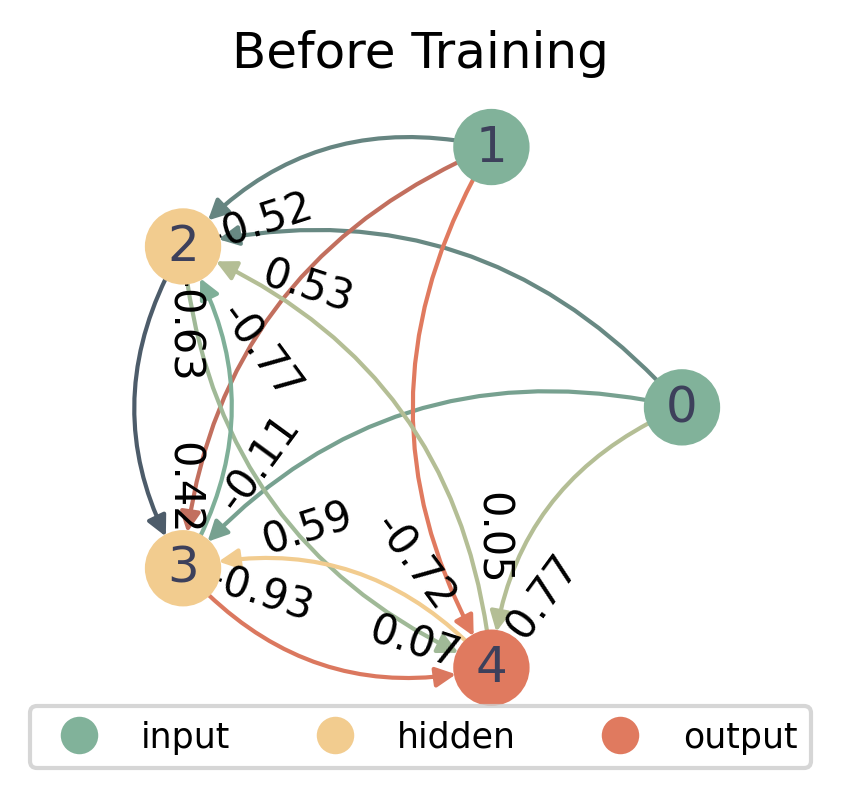}
    \includegraphics[width=0.4\linewidth]{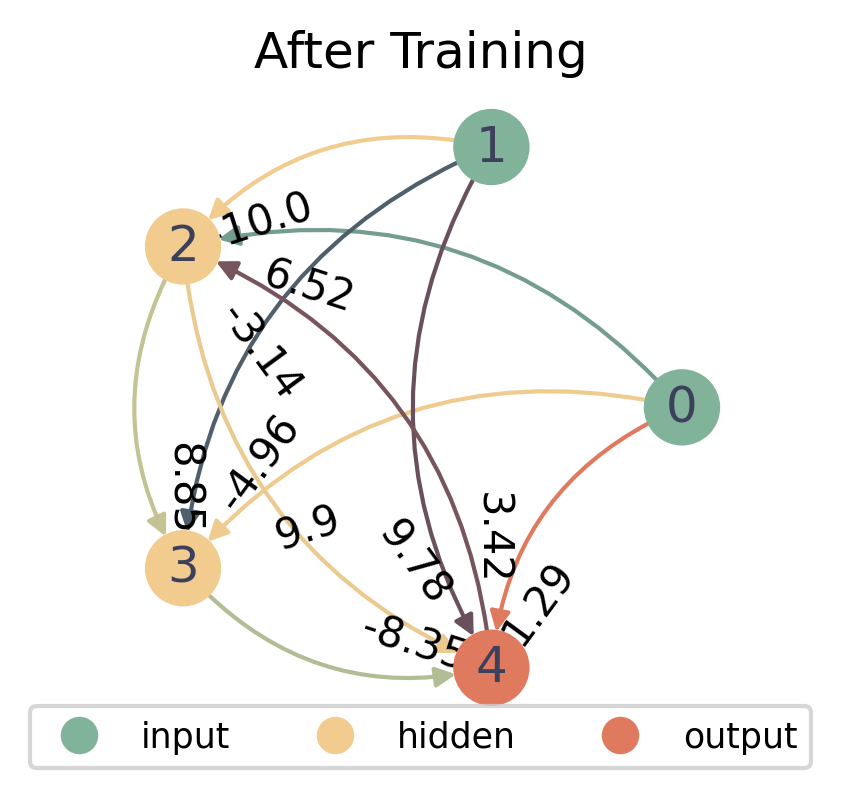}
    
    \caption{Example for an XOR gate; the edges $4 \rightarrow 3$ and $3 \rightarrow 2$ were removed during training.}

    \label{fig:example}
\end{figure}

\begin{figure}[t]
    \centering
    \includegraphics[width=1\linewidth]{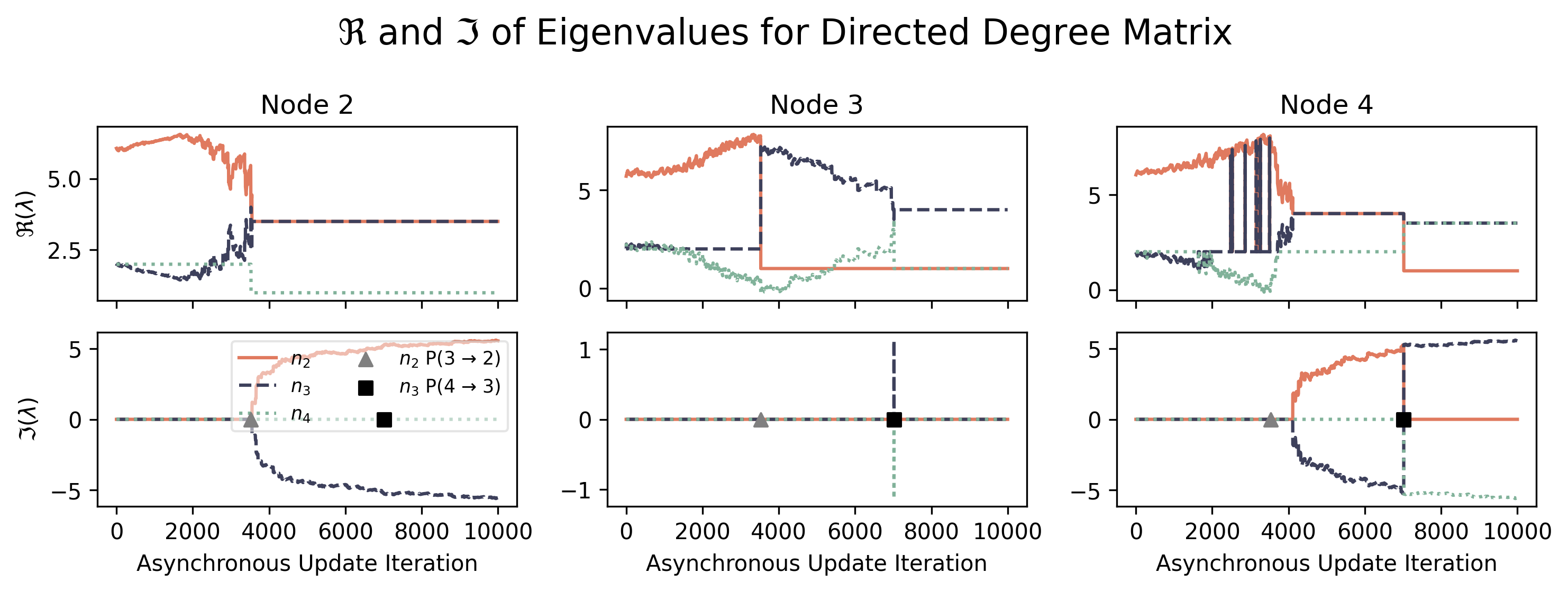}
    \caption{Development of the real and imaginary components of the eigenvalues shown for each node for the Degree mode $D$. Nodes $n_0$ and $n_1$ are excluded as they serve as placeholders for input values and do not participate in any behavior. Each color corresponds to the component value associated with a particular node. For clarity, consider node $n_2$: it prunes its input edge from $n_3$ at the moment this is detected (indicated by the legend $\triangle n_2 P(3 \rightarrow 2)$). This decoupling enhances the dynamic $n_2 \rightarrow n_3$, which cannot be pruned by $n_2$ since nodes are restricted to pruning only their input edges. Later in the optimization process, $n_3$ prunes its input edge from $n_4$, marked by $\square n_3 P(4 \rightarrow 3)$. The effect of this decoupling is visible in the neighborhood dynamics of $n_4$, where the imaginary component of $n_4$, initially zero, becomes coupled with $n_3$, in particular, the coupling appears in the edge $n_3 \rightarrow n_4$.}
    \label{fig:development}
\end{figure}

If two nodes have eigenvalues with opposite imaginary parts in their local Laplacians, it typically points to a form of structural or dynamic symmetry of coupled behaviors. Utilizing this insight in pruning allows for maintaining essential properties of the graph while reducing the amount of trainable parameters. Figure \ref{fig:example} shows an example before and after the dynamic pruning process for one of the networks evaluated in the experimental section. 
Figure \ref{fig:development} illustrates the evolution of the real and imaginary components of the eigenvalues computed for each node, along with the pruning operations executed.

\section{Experiments}
\subsection{Performance and Pruning Capacity}
\label{performance}
To assess the efficacy of the proposed approach, experiments were conducted on logic gates including $AND$, $OR$, and $XOR$. These experiments were carried out with the defined Degree matrix $D$ and $D_{W}$, and without pruning to establish a baseline. The experimental setup involved circuits with $2$ input nodes, $1$ output node, and a variable number of hidden nodes. The initial graph was set to be fully connected, except for the input nodes which act as placeholders and do not have input from any other node. The weights of the connections were randomly initialized with a uniform distribution between $-1$ and $1$.

During the experiments, each logic gate was subjected to $2000$ random input configurations. The nodes within the circuits executed their operations $10$ times per input, with the sequence of execution randomized for each input set. This update frequency ensured that each input configuration was processed multiple times to evaluate the robustness and efficiency of the proposed pruning mechanism. After the training process, the error was computed as the mean error of each input configuration of each gate.

\begin{figure}[t]
    \centering
    \includegraphics[width=0.7\linewidth]{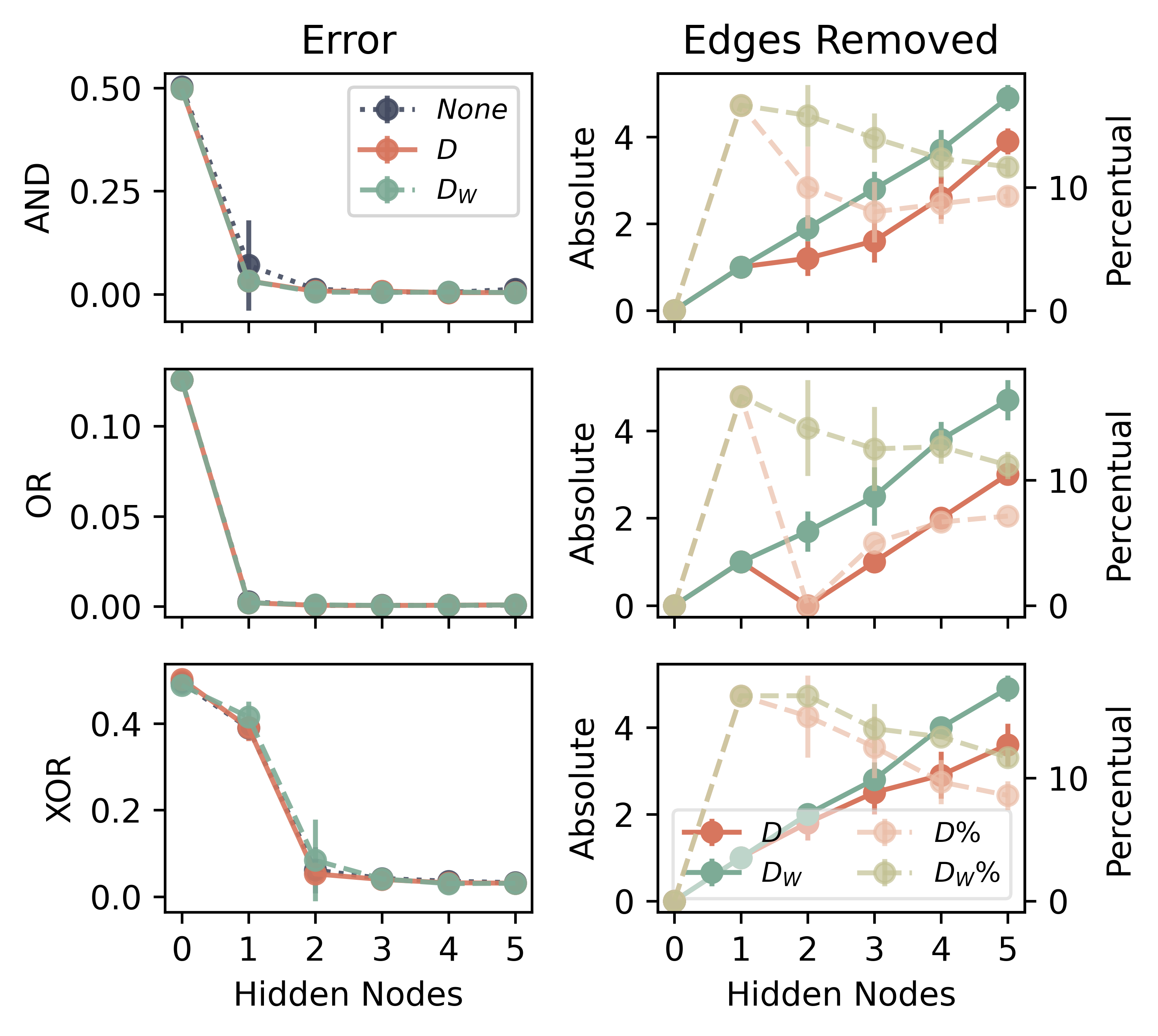}
    \caption{Experiment results: mean and standard deviation of $10$ runs per number of hidden nodes. The evaluated logic gates are shown with their respective after-training errors in the left column, for both Degree matrix modes directed $D$ and weighted directed $D_{W}$, $None$ indicates the unpruned network acting as a baseline for the error. The number of edges removed for different numbers of hidden nodes and both degree modes is shown in the right column in both absolute and percentual values.}
    \label{fig:results}
\end{figure}

Figure \ref{fig:results} illustrates the experimental results. A notable initial observation is that the pruning mechanism has minimal impact on the resultant error of the neural network, this means that the proposed method does not over-prune the network. Regarding the number of edges pruned it can be noticed that the degree mode $D_{W}$ has a consistently superior pruning capability than $D$.

\subsection{Pruning Timing}
\label{timing}
In network pruning, the timing of pruning refers to when the pruning is executed. According to \cite{chatzikonstantinou_recurrent_2021}, pruning can be categorized into three main types: pruning before training, pruning during training (with dynamic pruning falling under this category), and pruning after training. In dynamic pruning, early pruning during the initial stages of training involves removing parameters before the model fully converges. This approach can lead to a more efficient training process and may serve as a form of regularization, encouraging the model to learn robust features. However, early pruning also carries the risk of prematurely removing important parameters, which could destabilize the learning process and hinder the model's convergence.

\begin{figure}[t]
    \centering
    \includegraphics[width=0.7\linewidth]{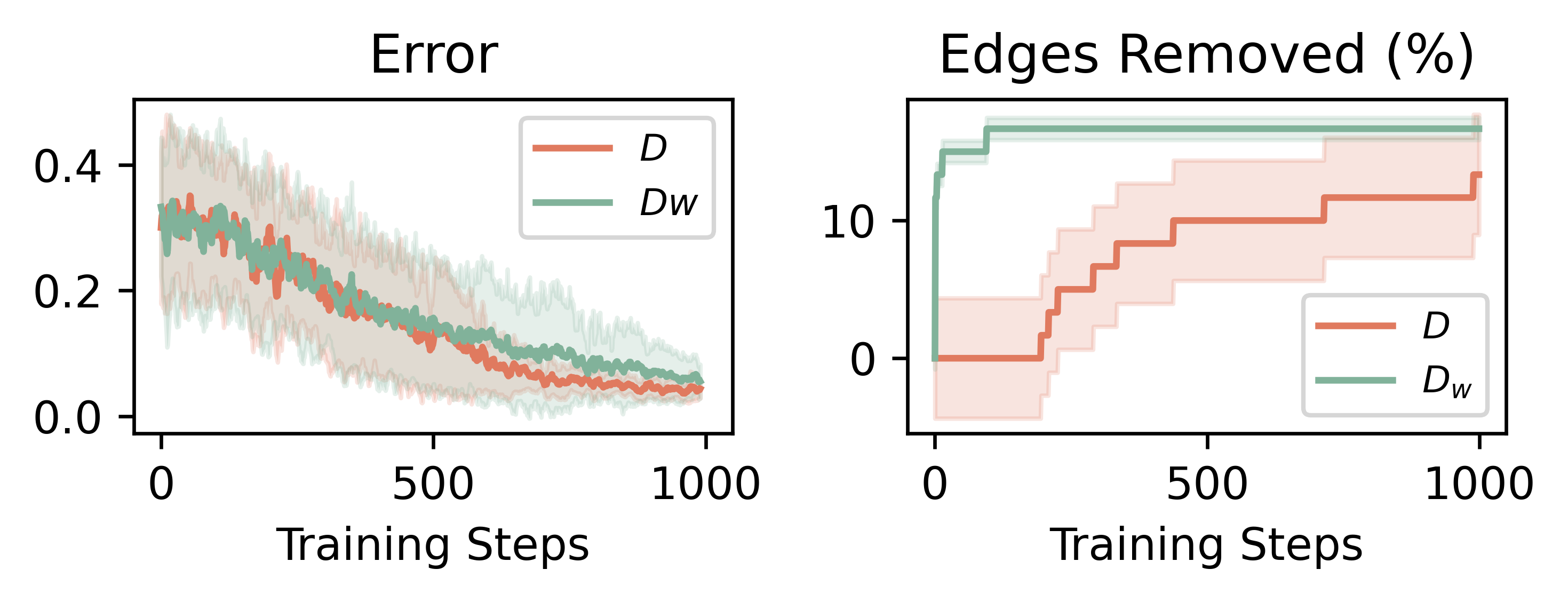}
    \caption{Mean and standard deviation of error and percentage of pruned edges during training for $5$ runs of a network with $2$ hidden nodes trained on the $XOR$ logic gate task.}
    \label{fig:timing}
\end{figure}

This experiment investigates the timing of pruning within the proposed methodology under both modes of the Degree matrix. Figure \ref{fig:timing} shows the result of the experiment, the initialization of the graph follows the same criteria as the previous experiment. Notably, in the $D_W$ mode, pruning occurs earlier compared to the $D$ mode. Although both models ultimately converge to similar values, the $D$ mode demonstrates a slightly faster convergence. This could be attributed to the larger parameter space in the $D$ mode, where it is easier to find a combination of parameters that successfully solves the task. In contrast, smaller models have a reduced number of parametric combinations, making the task of finding a suitable configuration more challenging.

This experiment underscores that varying the composition of the Degree matrix, and consequently the Laplacian matrix, can lead to distinct pruning behaviors. These differences can be strategically advantageous for specific tasks, suggesting that the choice of matrix composition should be carefully considered in the context of the desired outcomes. By tailoring the Degree and Laplacian matrix to align with particular objectives, we may enhance the performance and efficiency of pruning strategies in diverse applications.

\section{Discussion}

The experimental results from Section \ref{performance}, visualized in Fig. \ref{fig:results}, indicate that while the total number of pruned edges increases as more hidden nodes are added to the network, the proportion of pruned edges decreases with the growing number of hidden nodes. At first glance, this might suggest a limitation in the scalability of the methodology. However, this issue may not lie in the method itself, but rather in the pruning criterion. Recall that pruning occurs when two nodes have imaginary components of their eigenvalues that are equal and opposite. This method exclusively identifies binary couplings (i.e., direct interactions between two nodes), but as the network grows, higher-order couplings—interactions involving more than two nodes—become more prevalent. 

In networks with more hidden nodes, the likelihood of finding simple pairs of coupled nodes decreases, while more complex coupling patterns emerge, where oscillations are no longer confined to one-to-one relationships. This shift towards higher-order interactions requires more sophisticated methods to capture and prune these dynamics effectively. The current pruning decision process, while useful for testing the usefulness of spectral properties for pruning strategies, is a simplified approach. Future iterations should explore techniques that can detect and handle these multi-node interactions, potentially leveraging more advanced eigenvalue-based metrics or introducing new criteria that capture the full spectrum of node couplings. By addressing these complexities, we can improve the pruning process to better accommodate larger and more intricate networks, ensuring that we prune redundant connections without overlooking the subtle relationships that develop as the system scales.

In addition, in our current pruning strategy, we focus solely on the imaginary component of the eigenvalues, ignoring both real components (shown in Fig. \ref{fig:development}) and eigenvectors. While this approach provides a baseline for evaluating spectral properties, it overlooks the potential richness of the full spectrum. Future strategies could benefit from leveraging both real and imaginary components, capturing more intricate structural and dynamic properties of the graph. By refining these spectral insights, we can develop more sophisticated and efficient pruning techniques that adapt to the complexity of larger networks, ultimately improving both performance and scalability.

Regarding the composition of the Laplacian matrix, the results from both experiments reveal key insights into the impact of the degree modes ($D_{W}$ and $D$) on pruning effectiveness and network performance. The $D_{W}$ mode consistently prunes more edges than the $D$ mode, showcasing its superior pruning capability across both experiments. In terms of timing, pruning occurs earlier in the $D_{W}$ mode compared to the $D$ mode, yet both modes eventually converge to similar outcomes. The slight difference in convergence speed, with $D$ being faster, may result from the larger parameter space in the $D$ mode, which offers more combinations to solve the task. Conversely, the $D_{W}$ mode's reduced parameter space may present more difficulty in finding an optimal configuration, though it is still effective in the long term.

Different compositions of the Laplacian matrix can be used to adjust pruning timing, tailoring it to the needs of each task. For instance, a composition like $D_{W}$, which favors earlier pruning, could enhance efficiency in large, overparameterized networks by quickly reducing redundant connections. In contrast, the $D$ mode's slower pruning might be more suitable for tasks that require a larger parameter space for longer exploration. By modulating the balance between these modes, the pruning process can be adapted to optimize both performance and training efficiency.

\section{Conclusion}
The proposed approach leverages the spectral properties of local graph structures to devise a fully decentralized dynamic pruning strategy for asynchronous recurrent graph neural networks. This strategy was evaluated across different configurations of the local Laplacian matrix and network sizes, investigating how these configurations influence performance. The results highlight the promise of applying spectral theory to neural network pruning. Although the task appears straightforward, this research introduces innovative methods that settle the way for exploring new techniques to enhance structural efficiency among decentralized nodes while maintaining network performance.

This work excludes other significant spectral properties, such as eigenvectors and the real components of eigenvalues, which could potentially contribute to more refined pruning strategies. While the current results are promising, future research could benefit from incorporating these additional spectral features to further optimize and expand upon the use of spectral theory in pruning neural networks.

%
%


\bibliographystyle{spmpsci} 
\bibliography{references} 
\end{document}